\documentclass[runningheads]{llncs}

\usepackage{eccv}

\usepackage{eccvabbrv}

\usepackage{graphicx}
\usepackage{booktabs}

\usepackage{graphicx}
\usepackage{amsmath}
\usepackage{comment}
\usepackage{orcidlink} % for ORCID links
\usepackage{graphicx}
\usepackage{multirow}
\usepackage{rotating} % for rotating text
\usepackage{array} % for custom column alignment
\usepackage{adjustbox} % for adjusting box sizes

\usepackage[accsupp]{axessibility}  % Improves PDF readability for those with disabilities.

\usepackage{hyperref}

\DeclareUnicodeCharacter{2212}{-}

\usepackage{orcidlink}
\usepackage{marvosym}

\begin{document}

% ---------------------------------------------------------------
%
\title{Towards the Discovery of Down Syndrome Brain Biomarkers Using Generative Models}
\titlerunning{Discovery of Down Syndrome Brain Biomarkers via Generative Models}

\author{Jordi Malé\inst{1}\textsuperscript{(\Letter)}\orcidlink{0000-0003-4566-1921} \and
Juan Fortea\inst{2}\orcidlink{0000-0002-1340-638X} \and
Mateus Rozalem Aranha\inst{2}\orcidlink{0000-0001-9594-292X} \and
Yann Heuzé\inst{3}\orcidlink{0000-0002-0660-9613} \and
Neus Martínez-Abadías\inst{4}\orcidlink{0000-0003-3061-2123} \and
Xavier Sevillano\inst{1}\orcidlink{0000-0002-6209-3033}}

\authorrunning{J. Male et al.}


\institute{HER - Human-Environment Research Group, La Salle - URL, Barcelona, Spain\\
\email{jordi.male@salle.url.edu}\and
Sant Pau Memory Unit, Hospital de Sant Pau i la Santa Creu, Barcelona, Spain\\
%\email{\{jfortea,mrozalem\}@santpau.cat}
\and
Univ. Bordeaux, CNRS-PACEA, UMR 5199, Pessac, France\\
%\email{yann.heuze@u-bordeaux.fr}
\and
Departament de Biologia Evolutiva, Ecologia i Ciències Ambientals (BEECA), Facultat de Biologia, Universitat de Barcelona (UB),
Barcelona, Spain\\
%\email{neusmartinez@ub.edu}
}

\maketitle

\begin{abstract}
    Brain imaging has allowed neuroscientists to analyze brain morphology in genetic and neurodevelopmental disorders, such as Down syndrome, pinpointing regions of interest to unravel the neuroanatomical underpinnings of cognitive impairment and memory deficits. However, the connections between brain anatomy, cognitive performance and comorbidities like Alzheimer's disease are still poorly understood in the Down syndrome population. The latest advances in artificial intelligence constitute an opportunity for developing automatic tools to analyze large volumes of brain magnetic resonance imaging scans, overcoming the bottleneck of manual analysis. In this study, we propose the use of generative models for detecting brain alterations in people with Down syndrome affected by various degrees of neurodegeneration caused by Alzheimer's disease. To that end, we evaluate state-of-the-art brain anomaly detection models based on Variational Autoencoders and Diffusion Models, leveraging a proprietary dataset of brain magnetic resonance imaging scans. Following a comprehensive evaluation process, our study includes several key analyses. First, we conducted a qualitative evaluation by expert neuroradiologists. Second, we performed both quantitative and qualitative reconstruction fidelity studies for the generative models. Third, we carried out an ablation study to examine how the incorporation of histogram post-processing can enhance model performance. Finally, we executed a quantitative volumetric analysis of subcortical structures. Our findings indicate that some models effectively detect the primary alterations characterizing Down syndrome's brain anatomy, including a smaller cerebellum, enlarged ventricles, and cerebral cortex reduction, as well as the parietal lobe alterations caused by Alzheimer's disease. These results provide preliminary evidence supporting the automatic, data-driven discovery of brain biomarkers for Down syndrome and its associated comorbidities.
\end{abstract}

% TODO's
% 1 - Parlar de la variabilitat de les dades, on el diagnòstic no té perquè suposar alteracions severe
% 2 - Parlar de sagital vs axial, I que no tenim perquè trobar grans alteracions al MSP, sino que es imprescindible treballar en 3D

\section{Introduction}

\begin{figure}[h]
\centering
\includegraphics[width=1\textwidth]{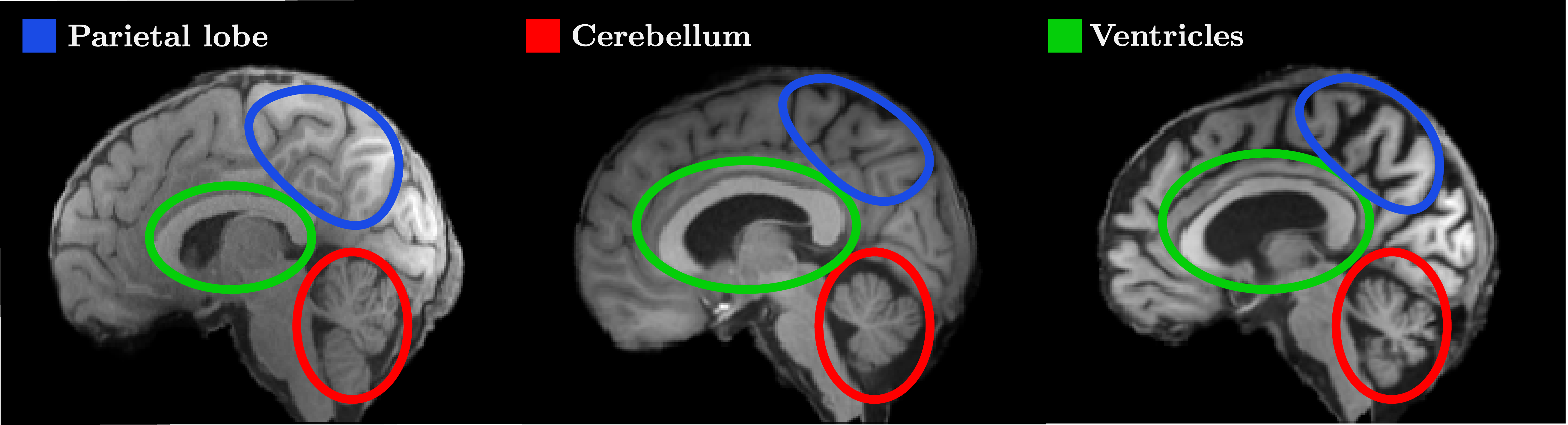}
\caption{Brain anatomy differences in the Mid Sagittal Plane (MSP) of three individuals: a control euploid (left), a person with DS with no signs of neurodegeneration (center), and a person with DS with advanced signs of dementia (right).} 
\label{fig:1}
\end{figure}

Brain imaging techniques, especially magnetic resonance imaging (MRI), are essential for studying the complex neurocognitive phenotype of Down syndrome (DS) and its related comorbidities \cite{Hamner2018PediatricBD}. Neuroimaging studies have revealed that persons with DS show a reduced overall brain volume from birth, with disproportionately smaller hippocampus and cerebellum, malformations in the corpus callosum and ventriculomegaly \cite{rodrigues}, along with premature brain aging marked by accelerated brain volume loss and progressive atrophy \cite{lao}. These structural brain changes are linked to cognitive and functional impairments from birth and early-onset dementia in adulthood, as individuals with Down syndrome older than 40 years old are at a high risk of developing Alzheimer's disease (AD) due to accelerated neurodegeneration \cite{rodrigues,fortea}. For illustration purposes, Figure \ref{fig:1} shows an example of the Mid Sagittal Plane (MSP) of the brain MRI scan of: \textit{i)} a person without DS (control euploid, or EU), \textit{ii)} a person with DS with no signs of neurodegeneration, and \textit{iii)} a person with DS with advanced signs of dementia. The image highlights the alterations caused by DS (enlarged ventricles and small cerebellum), and AD (deteriorated parietal lobe).

Despite advancements, the links between cognitive performance and brain anatomy, and between neuroinflammation and comorbidities like AD, remain unclear in DS. The rise of artificial intelligence (AI) offers the potential to develop automated techniques to assist clinicians in identifying disparities in brain anatomy between DS and EU subjects that lead to the discovery of diagnostic and prognostic brain biomarkers.

While supervised deep learning methods show great promise in brain analysis \cite{Male_2023_BMVC} and lesion detection \cite{supervised1,supervised2}, they require large, annotated MRI datasets, which are difficult to obtain due to privacy and ethical issues, and to the labor-intensive nature of annotation. Furthermore, they tend to generalize poorly beyond the learned labels \cite{supervised3}. To address these limitations, unsupervised, weakly-supervised and self-supervised AI techniques are gaining interest for both detecting brain alterations \cite{Tschuchnig_2022} and generating synthetic brain images \cite{pinaya2022brain}.

Autoencoders (AE) \cite{rumelhart_learning_1986} are one of the most promising unsupervised AI approaches for detecting brain alterations on MRI. When trained on large, diverse samples of brain MRI scans of a reference group, AEs learn a reliable model of that group's brain anatomy, which can then be used to detect alterations in new MRI scans \cite{reverse}. %Once trained, AEs can be used with generative AI techniques like Latent Diffusion Models (LDM) to produce high-quality synthetic 3D brain MRI scans \cite{synthesis}. These synthetic datasets have three main benefits: they enable training large AI models without size limitations, increase sample sizes for research, and facilitate the public sharing of MRI data without privacy concerns. 
Diffusion models (DM) have also been used for pixel-wise anomaly detection \cite{wolleb2022diffusion}, as autoencoder models are often complicated to train and have difficulties to preserve the finest details of the images. These weakly supervised models rely only on image-level labels for training (for instance, healthy vs. pathological), and can generate high-quality anomaly maps.

In this work, we apply three state-of-the-art anomaly detection techniques based on generative models --Vector Quantized Variational Autoencoders (VQ-VAE) \cite{vqvae}, Reverse Autoencoders (RAE) \cite{reverse}, and Denoising Diffusion Implicit Models (DDIM) \cite{wolleb2022diffusion}, which have been previously used to detect severe alterations such as tumors--  to detect subtle 2D brain alterations in MRI scans of individuals with DS. Our aim is to enhance the understanding of brain anatomical features in DS, paving the way for the discovery of biomarkers related to this syndrome and its associated comorbidities like AD. To achieve this, we leverage a proprietary dataset containing brain MRI scans of EU subjects and individuals with DS affected by varying degrees of neurodegeneration.

To validate the biological significance of the detected brain alterations and the effectiveness of the methods used, we conduct several evaluations: \textit{i)} a qualitative evaluation by expert neuroradiologists, \textit{ii)} a quantitative and qualitative evaluation of reconstruction fidelity on euploid control data, \textit{iii)} an ablation study on the impact of histogram post-processing on model performance, and \textit{iv)} a quantitative volumetric analysis using SynthSeg \cite{synthseg} to identify and quantify the regions most relevant to DS brain anatomy characterization.
%%CREC QUE AQUÍ CAL MARCAR PAQUET SOBRE ELS RESULTATS

The main contributions of this work are the following. First, to the best of our knowledge, this study pioneers the application of generative models for detecting brain alterations associated to different stages of neurodegeneration in DS. Second, we operate on mid-saggital plane views of the brain, in contrast to classic axial views used in anomaly detection, which entails a greater anatomical complexity. And third, we introduce a histogram matching post-processing step to enhance the quality of anomaly maps.

\section{Generative Models for DS Brain Alteration Discovery}

%\subsection{Reverse Autoencoder for DS Brain Alteration Discovery}
\subsection{Variational Autoencoders}

\begin{figure}[t]
\centering
\includegraphics[width=1\textwidth]{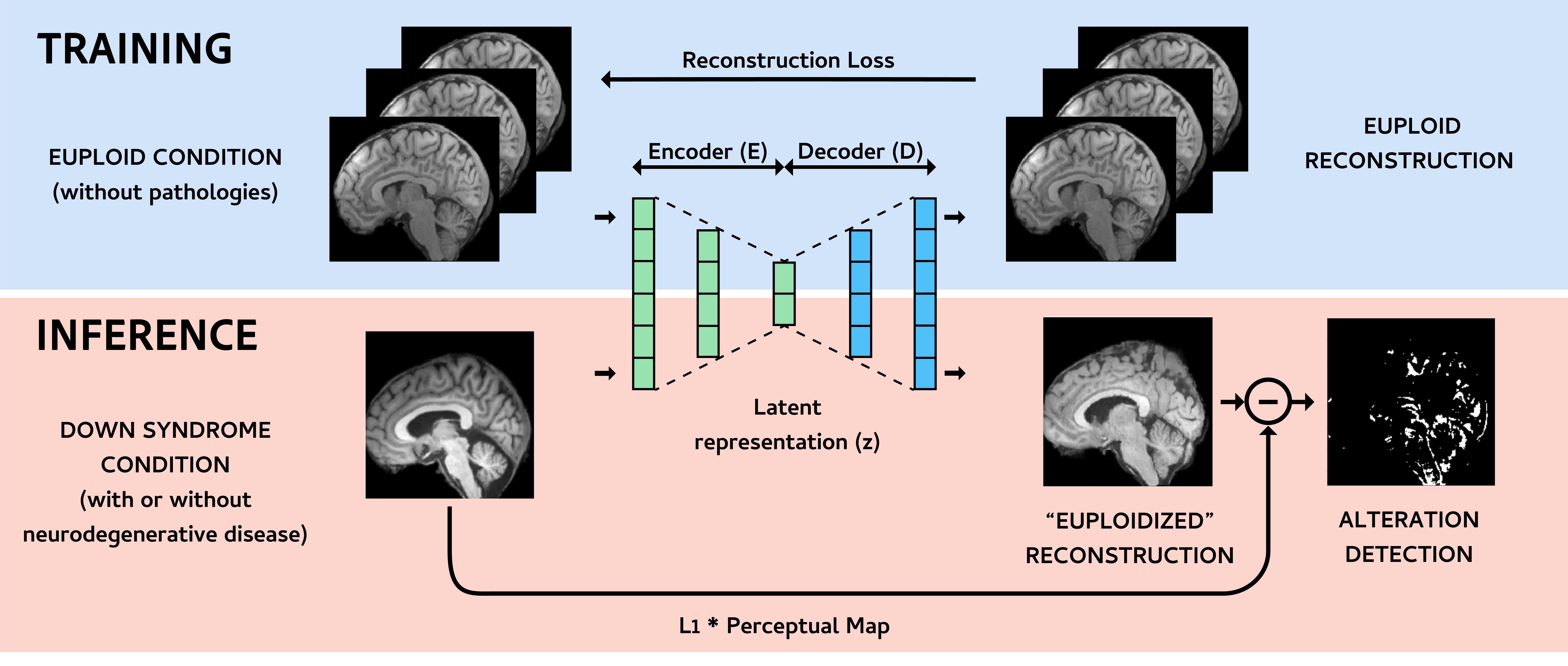}
\caption{Pipeline for DS brain alteration detection through Variational Autoencoders. L1 corresponds to Mean Absolute Loss.} \label{fig:2}
\end{figure}

The central concept behind the usage of variational autoencoders for unsupervised brain alteration discovery in DS relies on two core principles: first, the model is trained exclusively on EU subjects, learning the anatomy distribution of euploid brains; second, a brain MRI of a person with DS is fed into the trained network, resulting in a reconstruction of an ``euploidized'' version of the input. These reconstructions are then histogram matched to the input image to reduce intensity heterogeneity. To compute the anomaly map, the reconstructions are compared to the inputs to detect alterations, combining pixel-wise and perceptual differences \cite{perceptual}, along with histogram equalization, as proposed in \cite{reverse}. Figure \ref{fig:2} illustrates this procedure.

\subsubsection{Vector Quantized Variational Autoencoders (VQ-VAE)}\cite{vqvae} have demonstrated their ability to project high-resolution images into compressed latent representations \cite{vqvaeres}. They have been successfully used for unsupervised anomaly detection and other high-resolution reconstruction tasks, such as serving as an encoder for latent diffusion models.

VQ-VAEs enhance the representation learning capabilities of standard VAEs by introducing a discrete latent space. This discrete space provides more robust encoding and mitigates issues like posterior collapse, which are commonly observed in traditional VAEs.

In the VQ-VAE framework, the encoder \( E \) projects the input image \( \mathbf{x} \in \mathbb{R}^{H \times W \times D} \) into a latent representation space \( \mathbf{z} \in \mathbb{R}^{h \times w \times d \times n_z} \), where \( n_z \) is the dimensionality of the latent embedding vector. Each spatial code \( \mathbf{z}_{ijk} \in \mathbb{R}^{n_z} \) is then quantized to its nearest vector \( e_k \in \mathbb{R}^{n_z} \) from a codebook containing \( K \) vectors, resulting in the quantized latent representation \( \mathbf{z}_q \). The elements of the codebook are learned online, along with the other model parameters. Based on the quantized latent space, a decoder \( G \) reconstructs the input image \( \mathbf{\hat{x}} \in \mathbb{R}^{H \times W \times D} \).

The VQ-VAE framework employs two main types of losses: an \( L1 \) loss for reconstruction fidelity and a commitment loss to ensure the encoder outputs remain close to the codebook entries. This commitment loss prevents the codebook from becoming too sparsely populated and maintains robust quantization.

In our approach, we trained a VQ-VAE exclusively on euploid control data, focusing on minimizing the reconstruction error and maintaining high-quality latent representations through the combination of these losses.

\subsubsection{Reverse Autoencoders (RAE)} \cite{reverse} have demonstrated effectiveness in mitigating reconstruction errors by incorporating a reversed multi-scale embedding loss into the encoder and computing anomaly scores based on residual and perceptual differences. In the evaluated approach, the architecture was trained using the Evidence Lower Bound (ELBO) for both the encoder and decoder, along with a reversed embedding similarity coefficient to ensure that the input representations align with the embeddings of the generated reconstructions. Inspired by knowledge distillation methods \cite{Salehi}, this reverse comparison is conducted at multiple levels.

Thus, the autoencoder was trained to minimize the learning objectives of the encoder (Equation \ref{eq:encoder}) and the decoder (Equation \ref{eq:decoder}).

\begin{equation}
L_{E_\phi} (x, z) = \textit{ELBO}(x) - \frac{1}{\alpha} \left( \exp\left( \alpha \textit{ELBO}(D_\theta(z)) \right) + \lambda L_\textit{Reversed}(x) \right),
\label{eq:encoder}
\end{equation}

\begin{equation}
L_{D_\theta} (x, z) = \textit{ELBO}(x) + \gamma \textit{ELBO}(D_\theta(z))
\label{eq:decoder}
\end{equation}

where $L_{reversed}$ is the reconstruction embedding error, \(\lambda\) was empirically set to \(5 \times 10^{-3}\), and \(\gamma\) and \(\alpha\) are hyperparameters set to 0.5 \cite{reverse}.

\subsection{Diffusion Models}

\begin{figure}[t]
\centering
\includegraphics[width=1\textwidth]{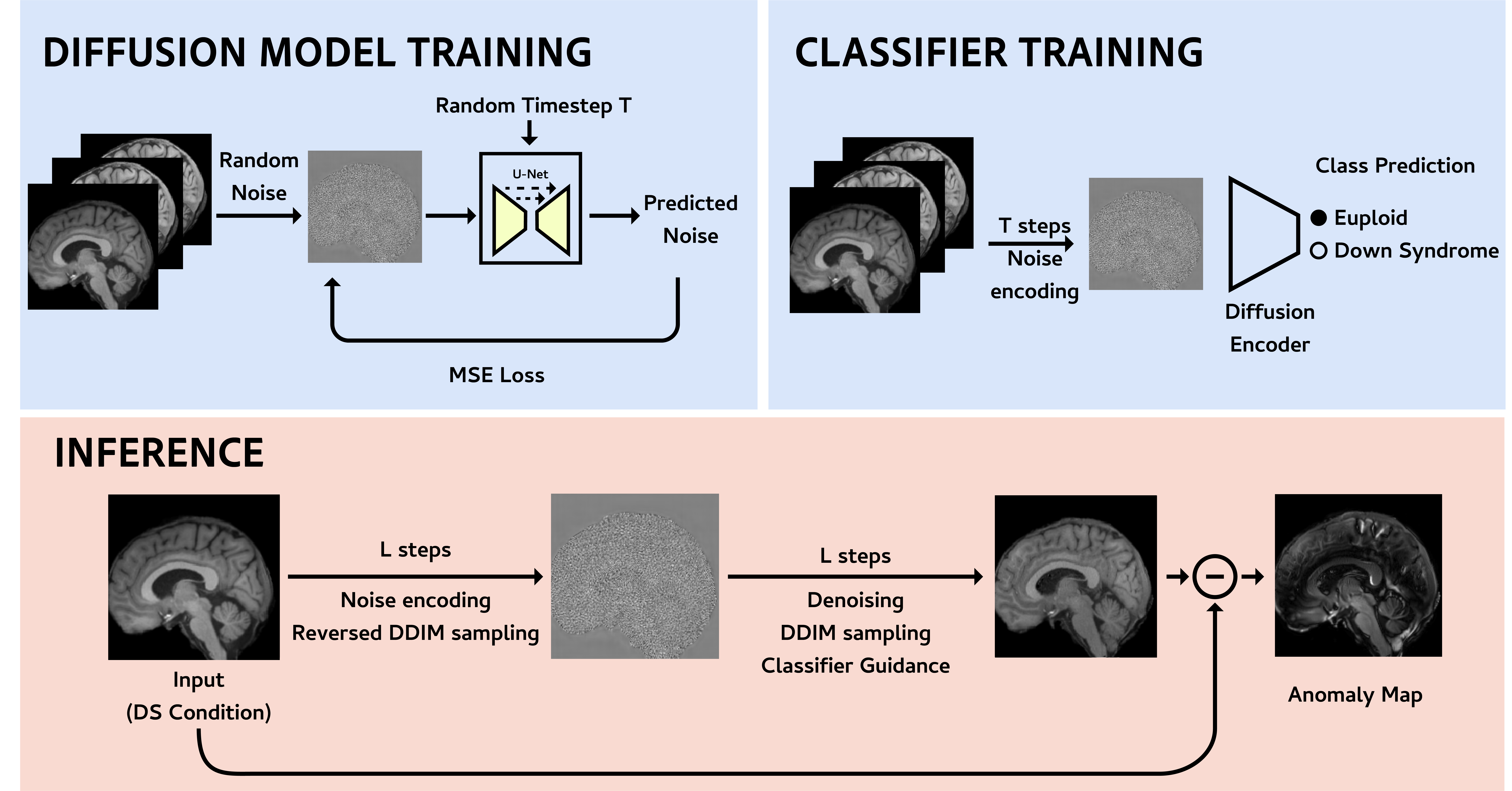}
\caption{Pipeline for DS brain alteration detection using Diffusion Models.} \label{fig:3}
\end{figure}

The fundamental principle of Diffusion Models (DM) for automatic DS brain alteration discovery is image-to-image translation \cite{diffusionReview}, which involves transforming a brain image of a person with DS into a brain image of a euploid person. For effective detection, it is essential that only the altered regions are modified, while the rest of the image remains unchanged. The anomaly map is then defined by the differences between the original and the translated images.

Diffusion Models (DM) have been recently applied to various tasks in medical imaging \cite{diffusionReview}, demonstrating excellent performance in anomaly detection, among other applications.% The fundamental principle of DM for automatic alteration detection is image-to-image translation. In the context of medical imaging, this involves transforming an image of a patient into one without any pathologies. %For effective anomaly detection, it is essential that only the pathological regions are altered while the rest of the image remains unchanged. The anomaly map is then defined by the differences between the original and the translated images.

In this work, we employed denoising diffusion implicit models (DDIM) and followed the methodology presented in \cite{wolleb2022diffusion}. %The general idea is that for an input image $\textit{x}$, we generate a series of noisy images $\{x_0, x_1, ..., x_T \}$ by adding small amounts of noise for many timesteps $T$. The noise level $t$ of an image $x_t$ is steadily increased from $0$ to $T$. A U-Net is trained to predict $x_{t - 1}$ from $x_t$, for any step $t \epsilon \{1, ..., T\}$. During training, we know the ground truth for $x_{t − 1}$, and the model is trained with an MSE loss. During evaluation, we start from $x_T \sim N (0, I)$ and predict $x_{t − 1}$ for $t \epsilon \{T, ..., 1\}$. With this iterative denoising process, we can generate a fake image $x_0$.
We trained a DDIM on a dataset containing images of EU and DS subjects. For evaluation, we defined a noise level \(L \in \{1, \ldots, T\}\) and a gradient scale \(s\). Given an input image \(x\), we encoded it to a noisy image \(x_L\) for \(t \in \{0, \ldots, L - 1\}\). This iterative noising process allowed us to induce anatomical information of the input image. Subsequently, the image was denoised for \(t \in \{L, \ldots, 1\}\) steps. We applied classifier guidance as introduced in \cite{diffgan} to steer the image generation toward the desired euploid class \(h\). For this purpose, we trained a classifier network \(C\) on the noisy images \(x_t\) for \(t \in \{1, \ldots, T\}\) to predict the class label of \(x\). During the denoising process, the scaled gradient \(s \nabla_{x_t} \log C(h|x_t, t)\) of the classifier was used to update \(\varepsilon_{\theta}(x_t, t)\). This iterative noising and denoising scheme is described in \cite{wolleb2022diffusion}. We generated an image \(x_0\) of the desired class \(h\) that retained the basic structure of \(x\). The anomaly map was then defined by the difference between \(x\) and \(x_0\). The choice of the noise level \(L\) and the gradient scale \(s\) was crucial for balancing detail-preserving image reconstruction and the freedom for translation to a euploid subject. Figure \ref{fig:3} presents a visual schematic of the training and inference processes of the diffusion model for discovering DS alterations.

\section{Experimental Setup}
\label{ExperimentalSetup}

Different datasets were used to train the proposed models. MSP images of 1,113 structural T1 MRI scans of euploid brains that are publicly available from the Human Connectome Project (HCP) were used to train the VQ-VAE and the RAE\footnote{Human Connectome Project, WU-Minn Consortium (Principal Investigators: David Van Essen and Kamil Ugurbil; 1U54MH091657) funded by the 16 NIH Institutes and Centers that support the NIH Blueprint for Neuroscience Research; and by the McDonnell Center for Systems Neuroscience at Washington University.}. This dataset was divided into 90\% for training (1000 subjects) and 10\% for evaluation (113 subjects). In contrast, the DDIM and the classifier were trained using the HCP dataset and 93 T1 MRI scans of subjects with DS obtained from a Philips 3 Tesla X Series Achieva scanner and provided by Hospital Sant Pau Memory Unit (Barcelona, Spain). The study was approved by the Sant Pau Hospital Research Ethics Committee, following the standards for medical research in humans recommended by the Declaration of Helsinki. All participants or their legally authorised representative gave written informed consent before enrolment. 

Models were evaluated using multiple datasets. For the EU control reconstruction fidelity evaluation, we tested the methods using three different datasets of control euploid brain structural T1 MRI scans: \textit{i)} 160 scans provided by Hospital Sant Pau Memory Unit (Barcelona, Spain), \textit{ii)} 338 scans from the OASIS-3 dataset \cite{oasis}, \textit{iii)} 506 scans from the IXI dataset \cite{IXI} and \textit{iv)} 113 scans from the HCP dataset (the 10\% of the dataset). For the DS alteration detection, models were evaluated using scans of individuals with DS, which were categorized into three subgroups based on their degree of neurodegeneration: \textit{i)} 61 DS subjects without signs of AD (no AD, mean age: 38 years), \textit{ii)} 11 DS subjects with incipient signs of dementia (prodromal AD, mean age: 52 years), and \textit{iii)} 12 DS subjects with advanced signs of dementia (AD, mean age: 52 years). Nine subjects were excluded from the evaluation due to uncertain neurodegeneration diagnosis.

All the MRI scans went through a preprocessing pipeline comprising bias correction, skull-stripping using Synthstrip \cite{Synthstrip}, and affine registration to the MNI-152 template. From each 3D volume, we extracted the MSP (the MRI slice that separates the brain into two almost-identical hemispheres) automatically through a multi-scale search algorithm that finds the plane that maximizes brain symmetry measured in terms of cross-correlation \cite{Ruppert}. This algorithm \cite{Male_2023_BMVC} employs a two-scale process: initially, it evaluates the cross-correlation of planes at a $\frac{1}{4}$ scale of the volume, then refines the search using the full volume. The multi-scale search ultimately identifies the MSP as the plane that maximizes brain symmetry on both sides. 

% ECCV - ADD THIS?

% ECCV - CHANGE THIS?
To improve the model's generalization to new data, data augmentation was employed during training through affine rotations of ±10°. Additionally, to address the significant class imbalance in the Down syndrome (DS) subject evaluations, all available DS mid-sagittal plane (MSP) images were used, along with the two adjacent sagittal slices on either side. This approach resulted in five slices per volume, yielding a total of 465 testing images (93 DS subjects with 5 slices per subject).

%For training our models, we used the extracted MSP along with the two consecutive sagittal slices on each side of it, resulting in a total of five slices per volume. Additionally, data augmentation with affine rotations of $\pm10^\circ$ was applied to enhance the models' generalization.

%\subsection{Implementation Details}

The VQ-VAE was implemented using the MONAI Generative framework\footnote{https://github.com/Project-MONAI/GenerativeModels}\cite{Monai}. The RAE was implemented based on the public official implementation\footnote{https://github.com/ci-ber/RA}). The DDIM was implemented using the MONAI framework and the public official implementation\footnote{https://gitlab.com/cian.unibas.ch/diffusion-anomaly} \cite{Monai}. Our code was developed using Python 3.9 and PyTorch, with experiments conducted on NVIDIA Tesla V100 GPUs.

The VQ-VAE was trained for 1000 epochs, and the RAE was trained for 2000 epochs, both using the Adam optimizer. The VQ-VAE was trained with a fixed learning rate of $5E-4$, while the RAE was trained with an initial learning rate of $5E-4$, but was gradually decreased to 1E-6 throughout training. The DM and the classifier were trained for 4000 epochs, also using the Adam optimizer, with a fixed learning rate of $2E-5$.

\section{Results}

%SI AJUNTEM FIGS 3 i 4

\subsection{Evaluation of euploid brain anatomy reconstruction} 
\label{sec:4.1}

The following experiments evaluate the reconstruction of EU brain anatomy in both qualitative and quantitative terms.

\begin{figure}[t]
\centering
\includegraphics[width=\textwidth, keepaspectratio]{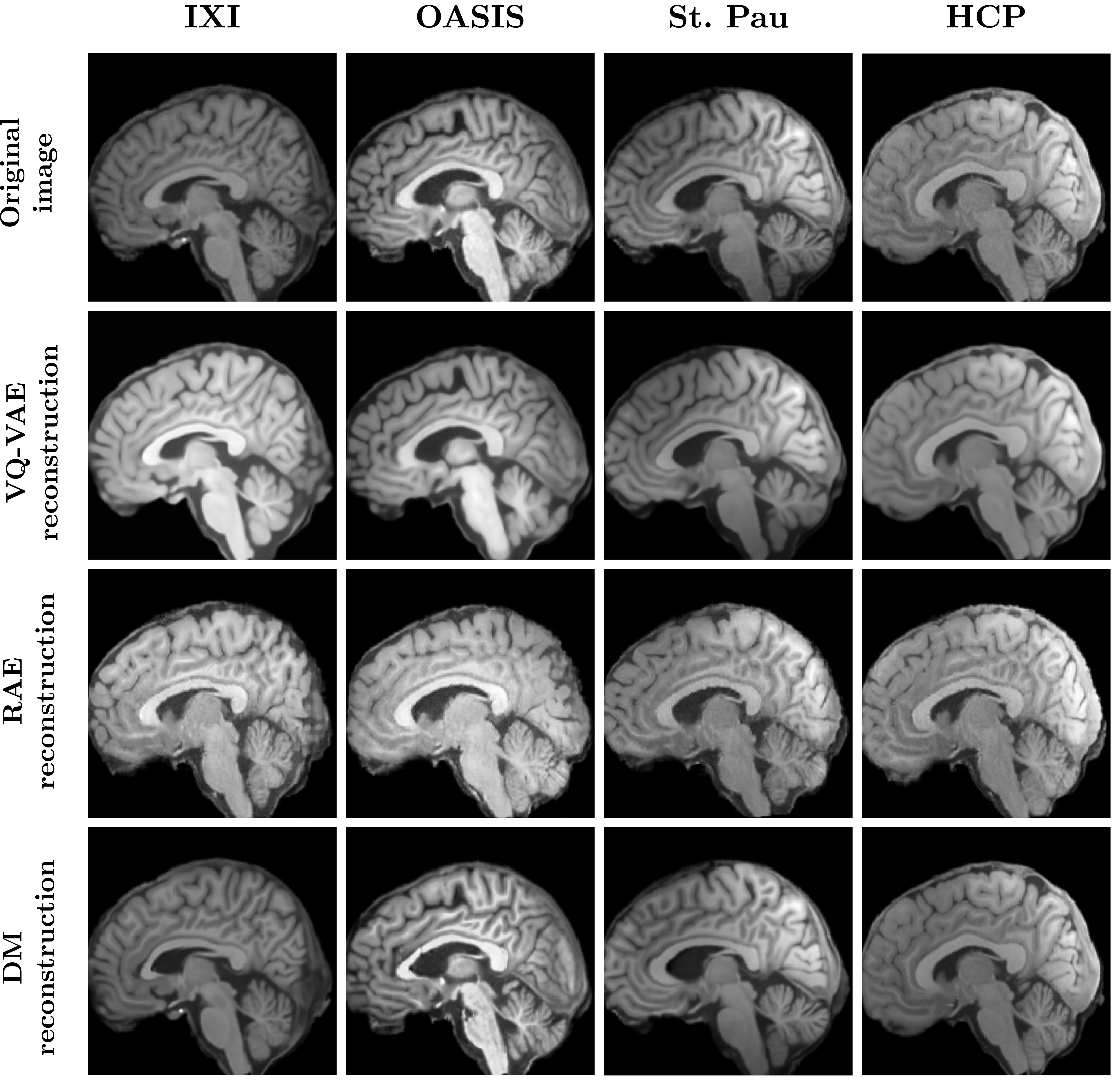}
\caption{Qualitative example of the models' performance on euploid subjects.} 
\label{fig:4}
\end{figure}

Figure \ref{fig:4} presents qualitative results of the trained models on the euploid datasets mentioned in Section \ref{ExperimentalSetup}. These results demonstrate that all models preserve the anatomy of EU subjects without altering any regions, as the models are trained on subjects from the same distribution. The diffusion model is evaluated by performing image-to-image translation on a EU subject, and it maintains the brain's anatomy throughout the noising-denoising process. 

% NEW - Discussion on metrics chosen

To quantitatively assess the fidelity of the reconstructions, Table \ref{tab:1} presents the Structural Similarity Index Measure (SSIM) and Mean Squared Error (MSE) of all models across different datasets. SSIM was chosen as it is sensitive to changes in image structure, making it suitable for evaluating the preservation of anatomical details in reconstructed brain images. MSE, on the other hand, offered a straightforward measure of pixel-wise differences, allowing for the detection of overall reconstruction accuracy. These complementary metrics together provide a robust assessment of model performance in maintaining both structural and quantitative integrity of the brain images. For this analysis, a subset of 10 subjects from each dataset (40 subjects in total) was used for inference due to time constraints. Additionally, we studied the effects of histogram matching to enhance the reconstruction's output and maintain the primary intensities of the input image. Histogram matching was performed using the model's output as the source histogram and the model's input as the target histogram. This technique improved both SSIM and MSE across all models, with notable improvements in the VQ-VAE and RAE models.

To thoroughly investigate how histogram matching affects model performance, we conducted ablation studies to evaluate its impact on reconstruction fidelity in the autoencoder-based models. This study is presented in Table \ref{tab:2}. We implemented four variants per model: \textit{i)} the original model without any modification (RAE, VQ-VAE), \textit{ii)} a new model trained with data matched to a template histogram (HRAE, HVQ-VAE), \textit{iii)} the original model with histogram matching applied from the model's output to the model's input (RAE + HIST M, VQ-VAE + HIST M), and \textit{iv)} a new model trained with data matched to a template histogram and with histogram matching from the model's output to the model's input (HRAE + HIST M, HVQ-VAE + HIST M). Results show that applying various histogram-based techniques can significantly improve the model's reconstruction fidelity, thereby enhancing performance in both SSIM and MSE.

\begin{table}[t]
\centering
\caption{Quantitative evaluation of the generative models on euploid subjects datasets, measuring the performance in terms of the SSIM and MSE metrics.}

% \caption{Quantitative evaluation of various generative architectures on euploid subject datasets, measuring performance using SSIM and MSE metrics.}
\label{tab:1}
\begin{tabular}{|c|c|c|c|c|c|c|c|c|}
\hline
\multirow{2}{*}{Method} & 
\multicolumn{2}{c|}{OASIS} & \multicolumn{2}{c|}{IXI} &\multicolumn{2}{c|}{St. Pau} & \multicolumn{2}{c|}{HCP}\\
\cline{2-9}
 & SSIM $\uparrow$ & MSE $\downarrow$ & SSIM $\uparrow$ & MSE $\downarrow$& SSIM $\uparrow$ & MSE $\downarrow$& SSIM $\uparrow$ & MSE $\downarrow$ \\
\hline
VQ-VAE & 0.748 & 0.006 & 0.724 & 0.003 & 0.725 & 0.002 & 0.755 & 0.0010 \\
\hline
VQ-VAE + HM & \textbf{0.921} & \textbf{0.001} & \textbf{0.905} & \textbf{0.002} & \textbf{0.919} & \textbf{0.001} & 0.938 & 0.0005 \\
\hline
RAE & 0.512 & 0.020 & 0.492 & 0.020 & 0.604 & 0.008 & 0.605 & 0.0080 \\
\hline
RAE + HM& 0.512 & 0.015 & 0.491 & 0.018 & 0.613 & 0.0060 & 0.614 & 0.0060 \\
\hline
DM & 0.769 & 0.009 & 0.785 & 0.006 & 0.854 & 0.005 & 0.930 & 0.0006 \\
\hline
DM + HM & 0.899 & 0.004 & 0.902 & 0.005 & 0.901 & 0.004 & \textbf{0.973} & \textbf{0.0003} \\
\hline
\end{tabular}
\end{table}

\begin{table}[h]
\centering
\caption{Ablation study evaluating the impact of histogram modifications on the performance of various UAD architectures across euploid datasets. Performance is measured using SSIM and MSE metrics.}
\label{tab:2}
\begin{tabular}{|c|c|c|c|c|c|c|c|c|}
\hline
\multirow{2}{*}{Method} & 
\multicolumn{2}{c|}{OASIS} & \multicolumn{2}{c|}{IXI} &\multicolumn{2}{c|}{St. Pau} & \multicolumn{2}{c|}{HCP}\\
\cline{2-9}
 & SSIM $\uparrow$ & MSE $\downarrow$ & SSIM $\uparrow$ & MSE $\downarrow$& SSIM $\uparrow$ & MSE $\downarrow$& SSIM $\uparrow$ & MSE $\downarrow$ \\

\hline
RAE & 0.513 & 0.0198 & 0.496 & 0.0220 & 0.520 & 0.0170 & 0.596 & 0.0080 \\\hline
HRAE & 0.525 & 0.0230 & 0.514 & 0.0250 & 0.542 & 0.0188 & 0.624 & 0.0066 \\\hline
RAE + HIST M & 0.512 & 0.0150 & 0.496 & 0.0166 & 0.520 & 0.0134 & 0.604 & 0.0066 \\\hline

HRAE + HIST M & 0.550 & 0.0113 & 0.523 & 0.0135 & 0.540 & 0.0108 & 0.622 & 0.0059 \\\hline

VQ-VAE & 0.756 & 0.0050 & 0.742 & 0.0040 & 0.743 & 0.0030 & 0.770 & 0.0010 \\\hline

HVQ-VAE & 0.759 & 0.0049 & 0.747 & 0.0038 & 0.773 & 0.0030 & 0.826 & 0.0010 \\\hline

VQ-VAE + HIST M & 0.907 & \textbf{0.0014} & \textbf{0.904} & 0.0015 & 0.907 & \textbf{0.0010} & 0.934 & \textbf{0.0005} \\\hline

HVQ-VAE + HIST M & \textbf{0.908} & 0.0015 & 0.894 & \textbf{0.0010} & \textbf{0.908} & \textbf{0.0010} & \textbf{0.938} & \textbf{0.0005} \\
\hline

\end{tabular}
\end{table}

\subsection{Evaluation of DS brain alteration detection} 

\begin{figure}[t]
\centering
\includegraphics[width=\textwidth, keepaspectratio]{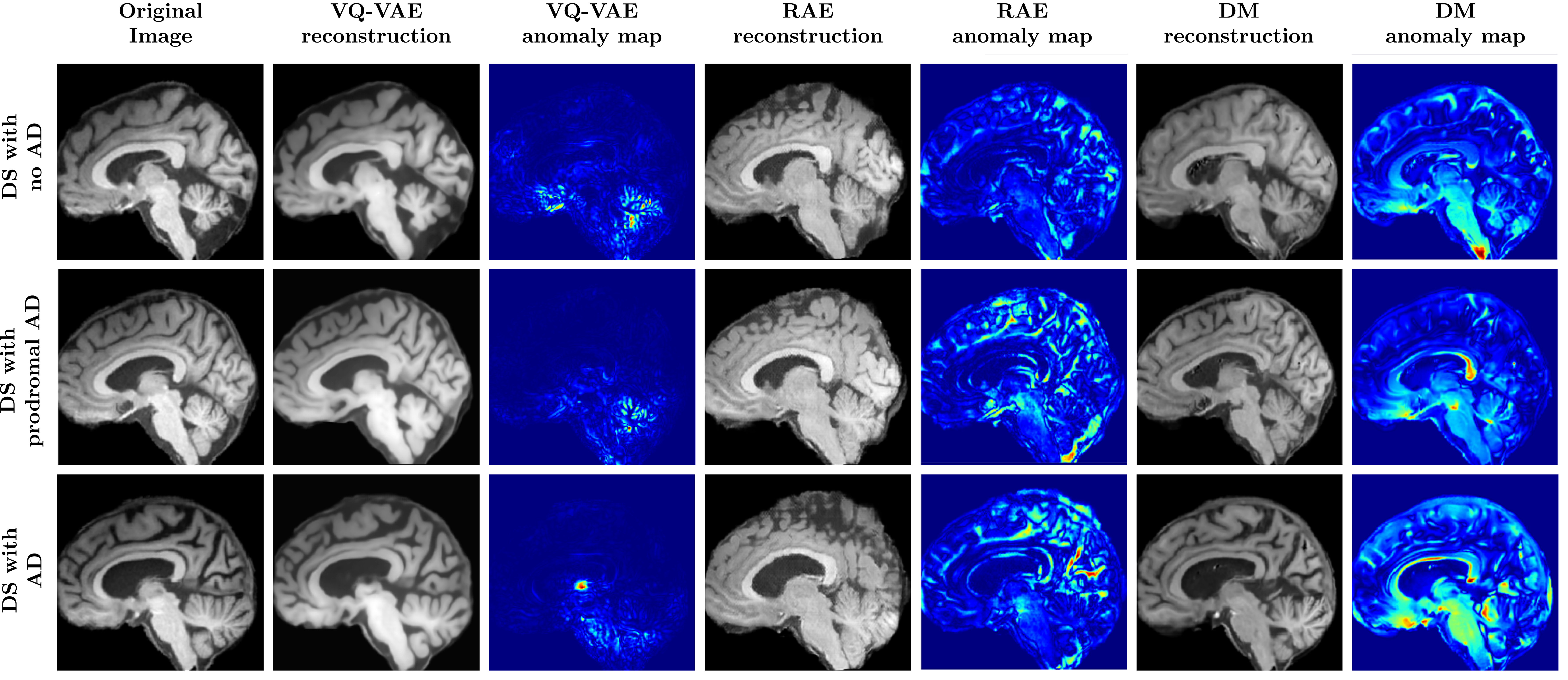}
\caption{Qualitative example of the models' performance across subjects with DS affected by different degrees of neurodegeneration: no AD (top row), prodromal AD (center row) and established AD (bottom row).} 
\label{fig:5}
\end{figure}

The following experiments evaluate the capability of the three generative models to detect brain alterations in individuals with DS affected by different degrees of neurodegeneration.

\begin{figure}[t]
\centering
\includegraphics[width=0.8\textwidth, keepaspectratio]{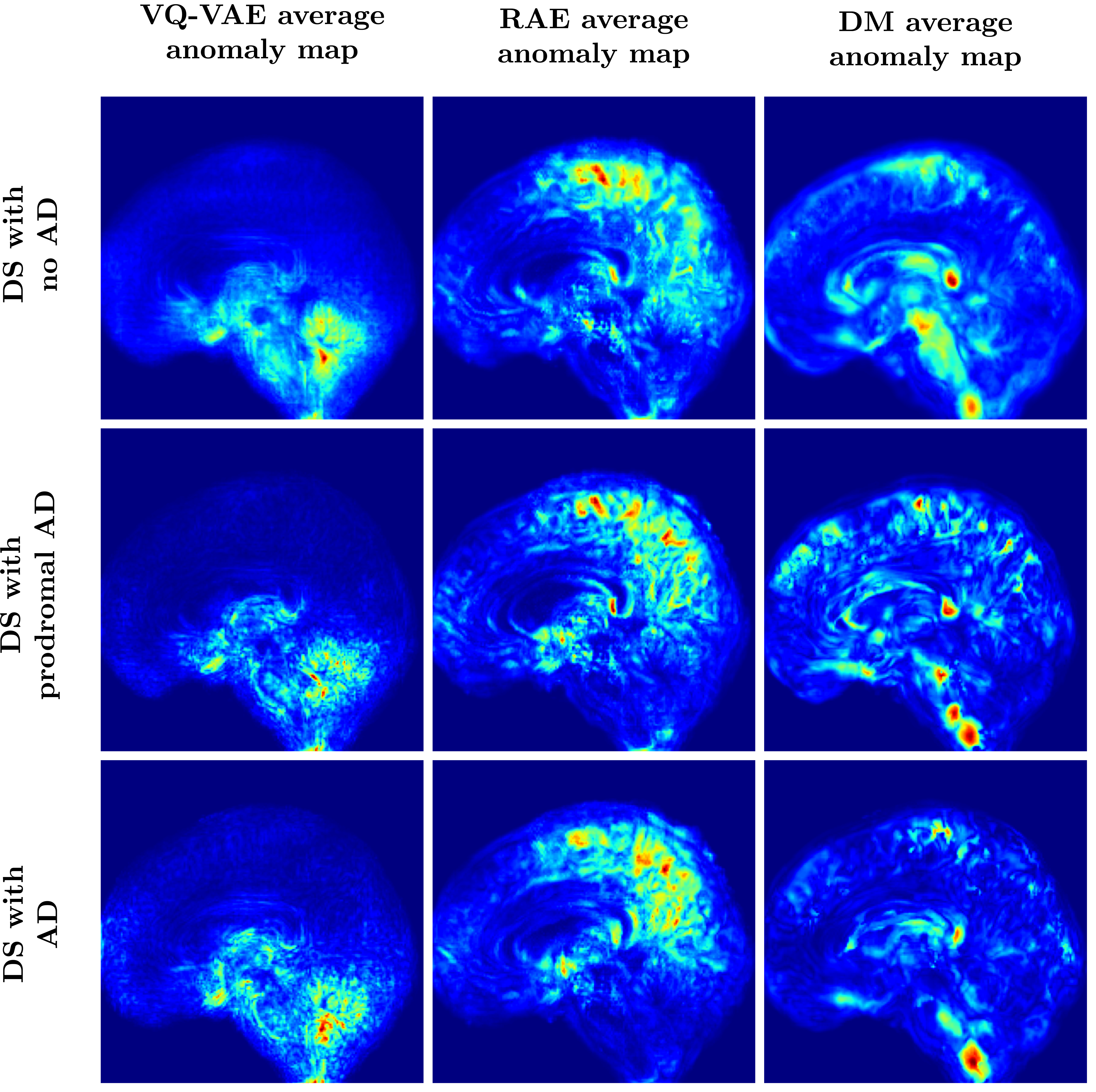}
\caption{Average anomaly maps across subjects with DS affected by different degrees of neurodegeneration: no AD (top row), prodromal AD (center row) and established AD (bottom row).} 
\label{fig:6}
\end{figure}

First, Figure \ref{fig:5} presents qualitative results of ``euploidized'' reconstructions and anomaly maps between EU and DS subjects. Each row of the figure displays the results for each DS subgroup (no AD, prodromal AD and AD). From left to right, we portray the original MRI scan of a particular subject, the reconstruction and anomaly map obtained through VQ-VAE, the reconstruction and anomaly map obtained through RAE, and the reconstruction and anomaly map obtained through DM. Then, Figure \ref{fig:6} presents the averaged anomaly maps obtained for each DS subgroup by RAE and DM, using all subjects in each subgroup.

% NEW - CANVIAT "a group of" per "two"
To qualitatively validate these results, two neurologists and radiologists with high expertise in DS from Sant Pau Hospital (Barcelona) analyzed the reconstructions and the anomaly maps. Their analysis confirmed that both the RAE and DM successfully generate ``euploidized'' reconstructions of the DS subjects, with increased cerebellum and decreased ventricles on all three DS groups, as these are two of the main features of DS brain anatomy. On the other hand, they stated VQ-VAE reconstructions did not produce ``euploidized'', as the model reconstructs almost the same input image. Moreover, the experts highlighted that DM reconstructions preserved better fine anatomical details of the brain, whereas RAE struggled with detecting fine-grain anomalies. This suggests the superior performance of DM in maintaining the integrity of brain structures, as confirmed in the evaluation in Section \ref{sec:4.1}.

However, the DM average anomaly maps highlight differences in the brain stem between EU and DS subjects. These differences are not true anatomical alterations; instead, they arise because the model struggles to reconstruct the region depending on the intensity of the input image, leading to high pixel-wise discrepancies that do not correspond to actual anatomical differences. Investigating how the methods in Table \ref{tab:2} can enhance the diffusion model's performance could significantly improve the quality of the average anomaly maps. This is particularly important because the DM reconstructions already outperform those of the other models, as shown in Figure \ref{fig:5}. 

More interestingly, the experts remarked that the evaluated models effectively detected brain alterations between DS subgroups caused by different degrees of neurodegeneration. Indeed, anomaly maps show the presence of alterations in the parietal lobes of prodromal AD and AD subgroups, which is one of the main brain areas affected by Alzheimer's disease \cite{AhulloFuster2022}.  

For further confirmation of these results, we validated the biological significance of these findings by conducting an automatic volumetric analysis of subcortical structures using Synthseg \cite{synthseg} on 84 T1 MRI scans of DS subjects (divided into three groups based on neurodegeneration diagnosis) and 77 T1 MRI scans of EU subjects, all obtained from the same scanner to ensure data consistency. Table \ref{tab:3} presents the average volumes (in mm$^3$) of the 12 subcortical structures that have higher relative volume differences between DS and EU subjects. For each DS subgroup, we present the relative volume difference (in \%) with respect to the EU subjects. 

We observe that the largest differences are congruent with the anomaly maps, focusing on the ventricles, cerebellum and hippocampus. This underscores the reliability and accuracy of the generative models in detecting and characterizing these alterations.

%Examining the mean anomalies detected, we observed that the volumes showing the most significant differences (Table \ref{table:1} corresponded to the same regions identified by our models. This congruence between the observed anomalies and the model predictions further validates the efficacy of our approaches. The regions with the most pronounced anomalies were consistent across the different groups, underscoring the reliability and accuracy of the unsupervised models in detecting and characterizing these alterations.

\begin{comment}
    \begin{figure}[t]
\centering
\includegraphics[width=\textwidth, keepaspectratio]{Images/combined.png}
\caption{Qualitative example of the models' performance across subjects with different conditions.} 
\label{fig:4}
\end{figure}
\end{comment}

%TODO formatejar els milers amb punts?

\begin{table}[t]
\caption{Top-12 average volumes (in mm$^3$) of subcortical regions showing the greatest differences between Down Syndrome (DS) and Euploid (EU) subjects. The DS subjects are further divided into subgroups based on the presence of Alzheimer's Disease (AD) neurodegeneration.}
\label{tab:3}

\centering
\begin{tabular}{|l |r |r r |r r |r r|}
\hline
\multirow{2}{*}{\textbf{Region}} & \multirow{2}{*}{\textbf{Euploid}} & \multicolumn{6}{c|}{\textbf{Down syndrome}} \\
\cline{3-8}
 &  & \multicolumn{2}{c}{No AD} & \multicolumn{2}{c}{Prodromal AD} & \multicolumn{2}{c|}{AD} \\
\hline
L inferior lateral ventricle & 895.7 & 1178.4 & (32\%) & 2155.8 & (141\%) & 2530.7 & (183\%) \\
L lateral ventricle & 16094.5 & 21115.9 & (31\%) & 32380.5 & (101\%) & 34557.2 & (115\%) \\
R inferior lateral ventricle & 863.5 & 1100.2 & (27\%) & 2190.0 & (154\%) & 2473.9 & (187\%) \\
R lateral ventricle & 14694.6 & 18468.8 & (26\%) & 26112.0 & (78\%) & 30709.1 & (109\%) \\
L cerebellum white matter & 25148.2 & 19576.1 & (-22\%) & 20434.9 & (-19\%) & 18534.4 & (-26\%) \\
R cerebellum white matter & 25288.2 & 19762.5 & (-22\%) & 20626.1 & (-18\%) & 18758.7 & (-26\%) \\
R putamen & 7274.4 & 8641.8 & (19\%) & 7956.2 & (9\%) & 6594.8 & (-9\%) \\
L putamen & 7499.4 & 8801.7 & (17\%) & 8095.3 & (8\%) & 6852.4 & (-9\%) \\
L cerebellum cortex & 67715.1 & 57297.9 & (-15\%) & 59742.7 & (-12\%) & 50625.0 & (-25\%) \\
R cerebellum cortex & 68025.5 & 58030.0 & (-15\%) & 58642.9 & (-14\%) & 51427.8 & (-24\%) \\
L hippocampus & 5868.5 & 5329.8 & (-9\%) & 4621.2 & (-21\%) & 3781.7 & (-36\%) \\
R hippocampus & 6036.3 & 5531.5 & (-8\%) & 4913.5 & (-19\%) & 3835.1 & (-37\%) \\
\hline
\end{tabular}
\end{table}

\section{Conclusions}

The primary objective of this study was to take the first steps towards automatic, unsupervised, and data-driven biomarker definition by comparing euploid subjects with individuals with Down syndrome. To that end, we have evaluated three state-of-the-art generative models for detecting anomalies in brain MRI scans, targeting individuals with Down syndrome at three different neurodegeneration stages. Moreover, a study on the impact of histogram processing on these models has been conducted, resulting in improved quality of the model's reconstructions.

Three generative model-based anomaly detection approaches were compared: a vector-quantized variational autoencoder, a reverse autoencoder, and a diffusion model. These models were evaluated using euploid data from different datasets to assess the fidelity of EU reconstructions both qualitatively and quantitatively. Additionally, they were tested on a proprietary dataset of brain MRI scans, with qualitative evaluations by expert neuroradiologists and quantitative assessments through a subcortical volumetric analysis to determine the biological significance of our findings.

Our experimental results demonstrate that models effectively detect regions with diagnostic potential by comparing reconstructions with the original images. Additionally, diffusion models surpass autoencoders in preserving fine anatomical details and identifying subtle anomalies. This highlights the clinical value of diffusion models, particularly in applications where maintaining the integrity of brain structures is crucial. Notably, different models detected different regions, suggesting that further research is needed to explore how these models can be used in tandem to detect subtle differences, as proposed in \cite{bercea2023reversing}. This could involve combining the strengths of each model, such as developing an ensemble approach that integrates the outputs of multiple models to improve detection accuracy and robustness.

Furthermore, our results indicate that leveraging traditional image processing techniques can significantly enhance the performance of generative models for image reconstruction. Employing various histogram-based methods has demonstrated improvements in both the reconstructions and anomaly maps.

% NEW - Ultima línia. Com ho veus aquí?

Future research will focus on several areas to enhance the models and broaden their applicability to various patient groups. First, we will explore new datasets with a greater number of Down syndrome subjects to improve the models' performance and generalization ability. Additionally, we will apply the models to identify biomarkers for other conditions, such as bipolar disorder and schizophrenia. As our results demonstrate, studying how traditional image processing techniques can complement generative models is also crucial. Furthermore, we will develop three-dimensional extensions of the models to define biomarkers based on overall volumes, which will be significantly more useful than the current approach that only utilizes the mid-sagittal plane. Finally, the study of the models using EU subjects has revealed some model-induced errors, which should be considered when interpreting differences in Down Syndrome patients.

\section*{Acknowledgements}
This work was partly supported by Agència de Gestió d’Ajuts Universitaris i de Recerca (AGAUR) of the Generalitat de Catalunya (2021 SGR01396, 2021 SGR00706), Agencia Española de Investigación (PID2020-113609RB-C21/AEI/ 10.13039/501100011033), the Fondation Jerome Lejeune under grant 2020b cycle-Project No.2001, and the Joan Oró grant (FI2024) from the DRU of the Generalitat de Catalunya and the European Social Fund (2024 FI-200014).

\bibliographystyle{splncs04}

\bibliography{references}
\end{document}